\documentclass[onecolumn,12pt,journal]{IEEEtran}
\ifCLASSINFOpdf
    \usepackage[pdftex]{graphicx}
\else
\fi
\usepackage{amsmath}
\usepackage{algorithmic}
\usepackage[ruled,vlined]{algorithm2e}

\usepackage[dvipsnames]{xcolor}

\usepackage{enumitem}

\begin{document}
%
\title{Process Knowledge Driven Change Point Detection for Automated Calibration of Discrete Event Simulation Models Using Machine Learning }
%
%
%

\author{Suleyman~Yildirim,
        Alper E.~Murat,
        ~Murat~Yildirim,
        ~Suzan~Arslanturk,
\thanks{Suleyman Yildirim, Alper E. Murat, and Murat Yildirim are with the Department
of Industrial and Systems Engineering, Wayne State University, Detroit,
MI, 48202 USA. Suzan Arslanturk is with the Department of Computer Science, Wayne State University, Detroit, MI, 48202 USA. \newline e-mail: suleymanyildirim@wayne.edu, amurat@wayne.edu, \newline murat@wayne.edu, suzan.arslanturk@wayne.edu}
}

%
%

\markboth{ 
}%
{Shell \MakeLowercase{\textit{et al.}}: Bare Demo of IEEEtran.cls for IEEE Journals}
%



\maketitle

\begin{abstract}

Initial development and subsequent calibration of discrete event simulation models for complex systems require accurate  identification of dynamically changing process characteristics. Existing data driven change point methods (DD-CPD) assume changes are extraneous to the system, thus cannot utilize available process knowledge. This work proposes a unified framework for process-driven multi-variate change point detection (PD-CPD) by combining change point detection models with machine learning and process-driven simulation modeling. The PD-CPD, after initializing with DD-CPD's change point(s), uses simulation models to generate  system  level  outputs  as  time-series  data  streams  which are  then  used  to  train  neural network  models  to  predict system  characteristics and change points.  The  accuracy of the predictive  models measures the likelihood that the actual process data conforms to the simulated change  points in system characteristics. PD-CPD iteratively optimizes change points by repeating simulation and predictive model building steps until the set of change point(s) with the maximum likelihood is identified. Using an emergency department case study, we show that PD-CPD significantly improves change point detection accuracy over DD-CPD estimates and is able to detect actual change points. 

\end{abstract}

%
\IEEEpeerreviewmaketitle

\emph{This work has been submitted to the IEEE for possible publication. Copyright may be transferred without notice, after which this version may no longer be accessible.}

\vspace{5mm}

Increasing complexity of modern systems require a new generation of simulation models that can accurately represent the time-varying system dynamics and provide decision support.  
In manufacturing and service systems, discrete event simulation (DES) models are extensively used to represent the discrete flow of materials, requests and customers in dynamic environments. 
Challenges associated with the complex nature of modern systems are being increasingly addressed by digital twin technologies \cite{RN36,RN37} that aim to create a one-to-one replica of the physical world using highly detailed simulation models. The resulting DES models reveal complex system relationships and unlock a significant potential for better prediction and decision making capabilities. However, these technologies also require a substantial upkeep as the underlying systems evolve. Dynamically changing characteristics of modern manufacturing and service systems necessitate adaptation of the DES models to represent the same system at different times. Hence, it becomes a fundamental research challenge to automate the detection and estimation of changes in system characteristics in order to maintain the validity of DES models. Conventional way of identifying these changes would be to use data-driven change point detection models (DD-CPD) \cite{RN38}. DD-CDP models rely on the system output(s) to detect when the system characteristics change. In reality, process dynamics in complex systems may alter (modulate and/or time-shift) how the changes in system characteristics manifest themselves in system outputs. 
Integrating process knowledge into change point detection models is thus critical to pinpoint the exact time of change in systems characteristics by accurately capturing how they relate to system outputs. This distinction proves to be increasingly critical as system complexities rise.

It is often difficult to predict the performance of modern systems through stylized models. 
Advances in data processing and computing capabilities enable powerful simulation models to incorporate and analyze these processes in detail. Conventional DES models assume a system with static and known characteristics. The main emphasis of these models are on modeling the stochastic behavior of different events in the system given system characteristics such as process flow, service and arrival distributions, and number of servers \cite{RN1}. However these characteristics change over time. While some of these changes are extraneous (i.e., arrival distributions) and can be detected only with DD-CPD methods, others such as process flow, resource levels and service time distributions are intrinsic to the system and their detection can greatly benefit from the process knowledge. In recent years, there has been an increasing interest in the automated discovery of process knowledge, with a specific focus on discovering \emph{control flow and process mapping} \cite{RN2,RN3}. Contemporary process mining literature prioritizes the discovery of control flow and extract the performance and resource patterns - i.e. resource levels and service distributions \cite{RN52}. Further, majority of the process mining approaches assume the process to be in steady state and other studies focus only on the \textit{concept drift} (changes in the process) in terms of control flow or resources to the extend they are detectable from the event logs \cite{bose2011handling, carmona2012online, martjushev2015change,bose2013dealing}. However, most of the dynamic changes in modern systems,  occur as a result of \emph{changes in performance and resource levels}, e.g., machine outage or degradation in a manufacturing setting and nurse schedule change or dynamic task assignment in healthcare systems. These resource level variations may not be detectable through the event logs due to, for example, lack of data collection or complex process inter-dependencies. Detection of these resource level variations and identifying the corresponding change points is a key challenge for migrating the static DES models to an adaptive and dynamic setting.





In this paper, we propose a process-driven change point detection (PD-CPD) method for complex systems using simulation models. The proposed PD-CPD method provides a unified framework that combines change point detection models with machine learning and process-driven simulation. The PD-CPD process is initialized using the DD-CPD estimates which are then refined through iterative simulation and machine learning  methodologies. In each iteration, given the change point(s) of the system characteristics, simulation models generate system level outputs as time-series data streams which are then used to train machine learning models to predict system characteristics. These machine learning models are utilized to estimate the likelihood that the data received from the physical world can point to the same change points in system characteristics. The process is repeated until the change point with the maximum likelihood is identified.  The unique contributions of this paper can be summarized as follows:
\begin{itemize}
\item We propose a change point detection method that focuses on the source of change - i.e. changes in system characteristics - as opposed to their manifestations in system outputs. Difference between PD-CPD and DD-CPD comes from the \emph{value of process information} that creates a mapping between system characteristics and outputs. This mapping cannot be incorporated within purely data-driven methods in many complex systems.
\item We develop an automated procedure to calibrate simulation models according to explicit changes that effect the underlying processes. For time-varying systems, this procedure enables the fully automated use of multiple DES models to represent the system at different times, rather than relying on an average analysis that cannot explicitly incorporate the dynamic system behavior.
\item We provide a method that uses machine learning to identify the fitness between the time series outputs from process-driven simulation model and physical world. The proposed method uses Nonlinear Autoregressive with External Input (NARX) Model that is more capable (e.g., handling non-linear and time-varying time series profiles) than alternatives such as linear parametric autoregressive, or autoregressive moving-average models \cite{RN35}.


\end{itemize}
In what follows, section 2 introduces the related work and surveys the development in change point detection and simulation research to place the contributions of this paper in context. In Section 3, we develop the unified methodology for PD-CPD and introduce its components. Section 4 presents a case study and demonstrates the effectiveness of PD-CPD. In sections 5 and 6, we conclude the paper with some limitations, future directions, and closing remarks.

\section{Related Work}
DES models systematize the operations of real-world system as discrete event sequences and commonly used to analyze dynamic and complex systems \cite{RN40}. DES models offer a flexible and dynamic system representation that are widely used for effective analysis of process-based systems in manufacturing, automotive, transportation, or healthcare industries \cite{RN40, RN39, RN56, RN57}. This paper aims to automate the detection and estimation of the changes in system characteristics to maintain the validity of DES models. Hence, it contributes to two independent literature streams: (1) automated discovery and calibration of DES models, and (2) change point analysis. 
\newline
\indent  
 Automated discovery and calibration of models has been studied by both the DES and process mining communities. In DES literature, many studies attempt to create an automated model discovery using formal model specifications for simulation models \cite{RN27, RN28, RN29, RN30}. More recently, several studies also propose data-driven model integration using external database and analysis techniques to create simulation models \cite{RN31, RN32}. In addition, some DES studies focus on model calibration and parameter update using data-driven approaches \cite{RN19, RN20, RN21}. Simulation model transformation and parameter calibration of DES formalism is discussed in different publications by Zeigler \cite{RN22, RN23}. Complexity of simulation model parameter calibration is also studied by Hofmann \cite{RN24}, and extensive development and evaluation of this procedure is analyzed by Park et al \cite{RN25}. Spear \cite{RN26} studied multivariate statistical analysis for calibration, uniqueness, and goodness of fit for high dimensional space and large simulation models of environmental systems. Model update and parameter calibration is also studied in process mining literature as concept drift \cite{bose2011handling, bose2013dealing}. Concept drift literature primarily focuses on detection of drifts on time \cite{RN44} and keeping predictive models up to date \cite{RN45}. Process mining studies present different methods to discover and understand business process model changes in a period \cite{bose2011handling, RN55}. 

\indent 
The proposed PD-CPD method is an offline supervised multivariate non-parametric change point detection approach using high-dimensional dependent time series data. Change point detection and estimation is a data-driven method to identify and diagnose abrupt changes in time series data. It is a signal processing tool not only commonly used in mathematics and statistics, but also in various fields such as machine learning, finance, economics, healthcare, engineering, etc. \cite{RN9,RN58,RN11,RN12}. Change point detection methods are traditionally classified as online (real-time setting) \cite{RN7, RN54} and offline (retrospective setting) detection \cite{RN8,RN13}. Most discrete event simulation models are executed offline, therefore our focus in this paper is on the multivariate offline change point detection approaches. Since there is no label to use in analysis, the change point detection problem is a typical unsupervised learning method. In the CPD literature, most methods consider one dimensional (univariate) data sets, and others use multi-dimensional (multivariate) data sets to detect changes. Change point detection methods for univariate time series has been studied for different domains in the literature \cite{RN11,RN15,RN16}. Since changes in a complex system may manifest across multiple system performance characteristics, it is important for the CPD method to handle  multi-dimensional data. In this paper, we also consider a nonparametric (distribution free) model with multivariate dataset, because we have no known distribution or a stable distribution in most simulation models. Matteson and James \cite{RN14} developed a nonparametric approach for multiple change point analysis of multivariate observations and compared with alternative methods. Other studies with nonparametric CPD approach in multivariate time series has been also analyzed in literature with different studies \cite{RN41, RN42, RN43}. 
\newline
\indent  
This paper mainly contributes to the automated DES model calibration literature by proposing a data-driven approach to detect and estimate the change points for resource parameter estimation. It also contributes to the nonparametric multivariate change point detection literature by leveraging the underlying process knowledge through simulation.

\section{Methodology}

This section develops the proposed PD-CPD method - a novel offline multi-dimensional change point detection approach driven by process knowledge and machine learning for automated change detection and calibration of DES models. Without loss of generality, we herein consider a class of periodic type changes, i.e. resource task assignment or shift change, in the system.

Figure 1 presents the flowchart of the PD-CPD methodology. Proposed PD-CPD method executes in four stages. First stage is the initialization stage and determines an initial estimate of the system's change point(s) by applying a multivariate nonparametric change point detection technique to the time-series dataset (for each realization). In the second stage, i.e. simulation stage, the method generates system outputs by simulating the DES model calibrated with the incumbent change points. 
The DES model is simulated for each realization of the system using the corresponding system inputs. Third stage is the predictive model training where we train a time series neural network model, i.e., Nonlinear Autoregressive with Exogenous Input (NARX) Model, for each realization. These neural networks are then used to predict the change points given the time-series datasets of the system's performance outputs. Last stage is the change point evaluation and perturbation stage. In this final stage, we evaluate the change point predictions obtained using the NARX models and update the change points using a perturbation based approach. We repeat this process with the perturbed change point(s) until change point(s) with  the  maximum  likelihood is identified. 

Inputs to the proposed PD-CPD method are multiple realizations of multivariate time-series data obtained from the system representing its performance outputs at different times. These multiple realizations correspond to unique output samples of the system in which the changes occur at the same change points, e.g., multiple day outputs of a system where changes are repeated daily. In addition, we process the raw data to extract parameters of a simulation model of the system. Outputs of the simulation model are also processed using a snapshot strategy to collect multi-variate time-series performance output data (features) i.e. number in system, number in queue, utilization etc. matching the time points of the actual system's time-series data. 

In what follows, we introduce the methods required for the proposed framework, and integrate them into a unified solution algorithm for process-driven change point prediction.

\begin{figure}[h]
    \centering
    \includegraphics[scale=0.9]{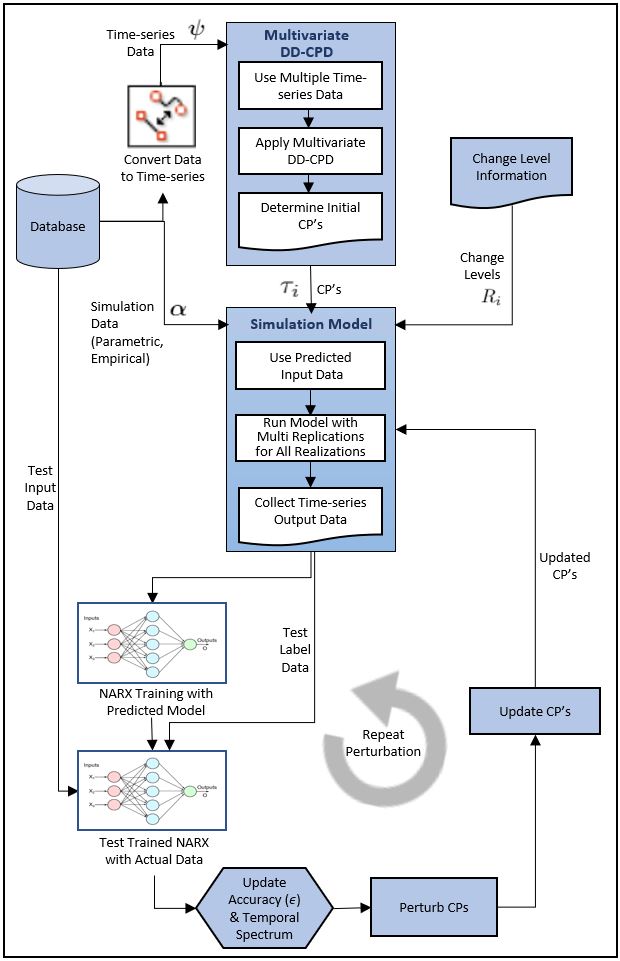}
    \caption{Process Knowledge Driven Change Point Detection (PD-CPD) Model Framework.}
    \label{fig_1}
\end{figure}

\subsection{Data Driven Change Point Estimation}
\label{Sec:DataDriven}
Multivariate non-parametric CPD is the first step of our methodology to detect data-based abrupt changes using real system data. This approach provides initial CPs using raw data, without using any process knowledge. We use standard offline change point detection methods, see references \cite{RN13}. There are multiple methods used for offline non-parametric change point detection such as non-parametric maximum likelihood, rank-based detection, kernel-based detection, and probabilistic methods. Change point detection method for a single change point $\tau$ can be expressed in general form as follows:
\begin{equation}
    \tau = \arg \min \bigg\{ Z(t)=\sum_{i=1}^{t-1}\varphi(x_i;\;\theta([x_1,\; \cdots \;x_{t-1}])+\sum_{i=t}^{n}\varphi x_i;\:\theta([x_t,\; \cdots \;x_n])\bigg\}
\end{equation}
where $\phi$ and $\theta$ define the section empirical estimate, and the deviation measurement, respectively. The formulation can use different types of statistics such as standard deviation, root mean squared, linear etc.

Since, we have $N$ number of change points in the case, then CPD function minimizes
\begin{align}
     Z(N)=\sum_{j=0}^{N-1}\sum_{i=t_j}^{t_{j+1}-1}\varphi(x_i;\:\theta([\theta_{t_j}\:...\:\theta_{t_{j+1}-1}])) + \beta N
\end{align}
where $t_0$ and $t_N$ are the first and the last sample of time series data, and constant number $\beta$ represents a fixed penalty added for each change point \cite{RN33, RN34}.

Unique to our application, the multi-variate raw dataset may represent multiple realizations of the system's performance data (i.e., periodic changes in the underlying system characteristics). For instance, the raw time-series dataset consists of the daily performance outputs of a system over the course of multiple days, i.e. realizations. Hence, applying a DD-CPD method to each realization could lead to a distinct set of change point(s). These distinct CPs implied by multiple realizations can be conciliated by selecting a representative CP set based on median or mean across realizations. The proposed PD-CPD method is agnostic to the DD-CPD method used to initialize and the initialization method of multiple CPs.

\subsection{Process-Driven Change Point Prediction}
Most process-driven systems have resource level changes in the short term (e.g. shift changes during a day or ad-hoc task assignment due to congestion) and long term (e.g. every 2-3 months or seasonal changes). These changes can be classified as one-of-a-kind (e.g., ad-hoc task assignment in service processes or machine degradation in manufacturing) or periodic (shift changes). In the case of periodic change points, the real system generates multiple realizations of time series data. The historical data from the system (e.g., event logs) is processed to generate features of the system performance in the form of time series data such as average number of entities in system, average number in queue, utilization of a process etc. using a sliding time window approach. Figure 2 illustrates the multi-dimensional time-series feature data for a single change point for a single realization of the system. 

\begin{figure}[h]
    \centering
    \includegraphics[scale=0.45]{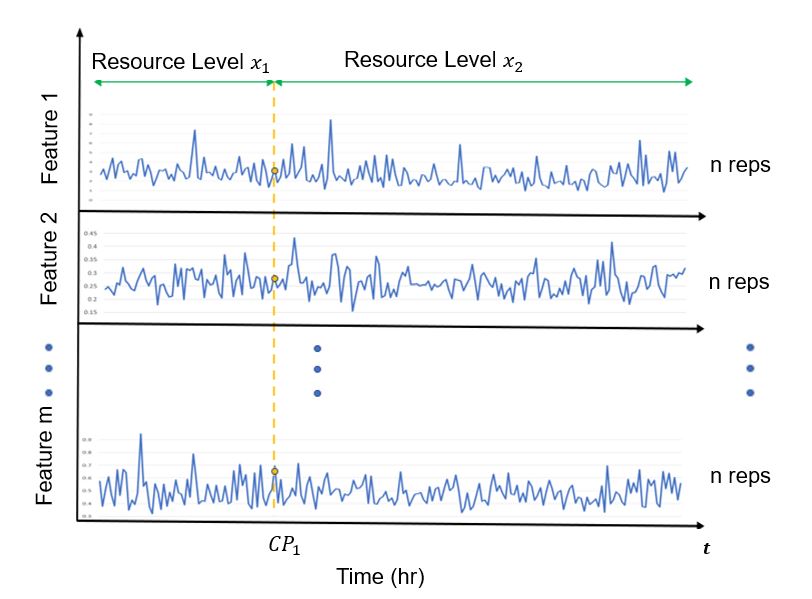}
    \caption{Multivariate time-series features of a process with a single change point.}
    \label{fig_2}
\end{figure}

Determination of which features to used is also a decision to be made as part of the proposed approach. For instance, if the goal is to estimate resource shift changes of a process step, then clearly utilization of that specific process as well as queues prior to the execution of that step should be included in the feature set. While some features are more important than others, the importance determination of features is left to the predictive model building step where less important features can be eliminated or new features can be extracted in a supervised manner \cite{li2018feature}. Such feature selection approaches are beyond the scope of our contribution. In addition, time windowing used to create time-series features of system performance is important. Clearly, these feature sets can be extracted at different levels of resolution by choosing the windowing function and temporal width. Choice of short width promotes greater sensitivity with respect to changes, but by increasing the size of time-series dataset it decreases the performance of the predictive modeling due to the system's intrinsic variability during periods with no changes. In comparison longer width windowing gives a better predictive performance in stationary periods, but reduces sensitivity to changes. Next, we discuss simulation model, NARX neural network, and change point perturbation steps of the PD-CPD.

\subsubsection{DES Simulation Modeling - Mapping System Characteristics to Performance Outputs} \label{Sec:SimModel}
DES models allow us to change the system characteristics and observe how those changes are manifested in the system outputs, therefore, it provides a powerful tool to map system characteristics changes with outputs. In implementing the PD-CPD's second stage (simulation stage), we assume that a DES model of the system is available or can be constructed. Further, we assume that the change points are characterized such that which process step they affect and their number are known except their timing.  
For instance, in our case study of emergency room modeling, we consider dynamic shift changes on resource (staffing) levels. We further assume that all system changes are same type (periodic or one-of-a-kind) and share the same periodicity (if periodic). In the case of periodic changes, we repeat the DES simulation runs for each realization of the historical data using the same flow unit attributes and arrival distributions but allow for variability through the processing times. Process duration distributions are estimated by distribution fitting using all realizations' data. DES models are configured to record high fidelity system performance data (e.g., queue lengths and wait times, total time in system). Each DES model corresponding to a realization is replicated multiple times with different random seeds and their outputs are processed through the same sliding time windowing approach used for the actual system's dataset to obtain multi-dimensional feature sets. All DES model runs are executed with the same incumbent change points (timing and levels).


\subsubsection{ Change Point Prediction and Evaluation - Nonlinear Autoregressive Exogenous Input (NARX) Neural Network Model} \label{Sec:NARX}
Given the simulated system's performance feature sets,  we train a unique neural-network  model (corresponding to each realization) which predicts temporal resource levels given the time-series features. Next we test the prediction accuracy of resource levels using actual system's features as inputs and simulated changes in resource levels as labels. For predictive modelling, we use Nonlinear Autoregressive Exogenous input (NARX) model, which is a robust class of dynamic recurrent neural network model suitable for nonlinear systems and time series and is non-parametric.  NARX predicts future values of a time series $y(t)$ from past values of that time series and past values of a second time series $u(t)$. The NARX neural network output can be mathematically expressed as follows:
\begin{align}
    y(t)=f[& y(t-1),y(t-2),...,y(t-d), u(t-1),...,u(t-d)] + \epsilon(t)
\end{align}
where $f$ is a nonlinear function that describes the system behaviour and $\epsilon(t)$ is the approximation error. The NARX neural network framework is illustrated in Figure 3.

\begin{figure}[h]
    \centering
    \includegraphics[scale=0.55]{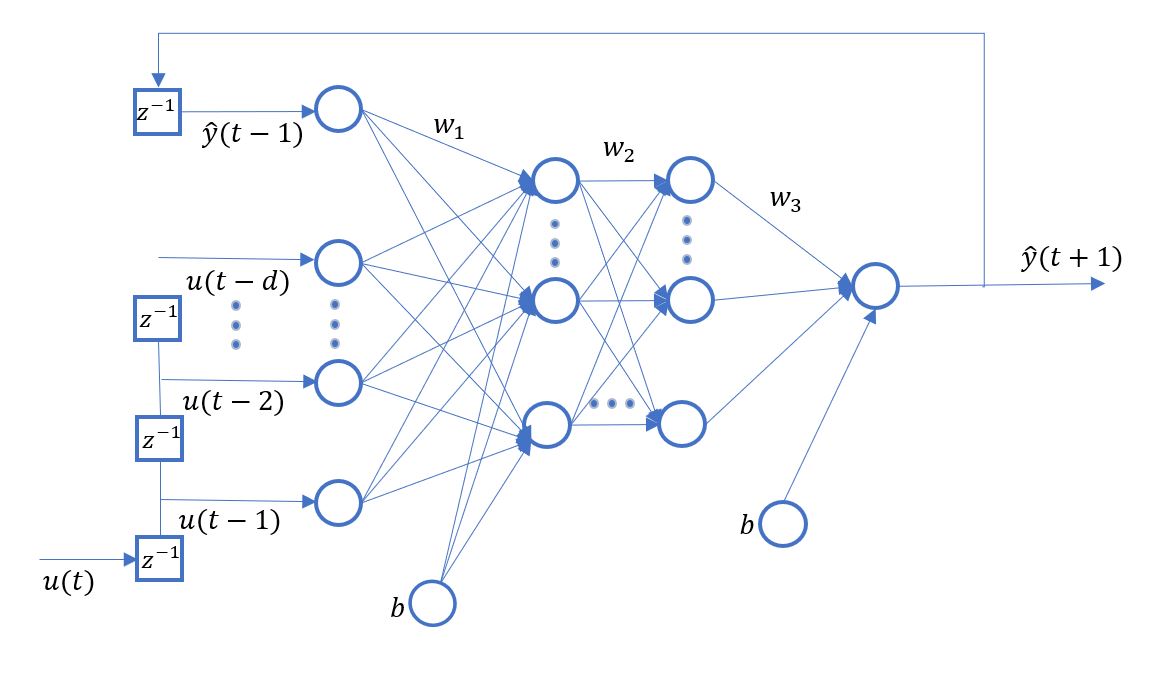}
    \caption{NARX Neural Network Framework with Delayed Inputs and Outputs.}
    \label{fig_3}
\end{figure}

In our approach, NARX neural network model predicts the temporal resource levels $y(t)$ using $d$ preceding resource levels and time-series performance features $u(\tau)$ where $t-d\leq\tau\leq t-1$. In training the NARX neural network, we use the the input $m$ dimensional feature data from the multiple DES replications (of a realization) and the simulated resource levels (as per the incumbent change). Note that the number of DES model replications (for a given realization) determines the size of data available for training and validation of the NARX model. For validation of the NARX model, we use k-folds in training and validation approach. To determine the optimal NARX model, we apply the common trial-and-error method to specify the number of hidden neurons and training function presented by Maier and Dandy \cite{RN48}. However,  the number of neurons, training functions, and other parameters can change depending on different case study. We explained all the details of our model parameters and input data in the case study section. The prediction performance of trained models were compared using mean squared error (MSE), accuracy measures. The lowest MSE or highest accuracy score provides the best prediction performance for that specific model. Once trained, each NARX model is then input with the features obtained from the actual system's realization to make resource level predictions. These resource level predictions are then discretized by a simple rounding procedure and compared with the simulated resource levels at a selected temporal region. The reason for not comparing the predicted resource levels with those simulated in the full temporal spectrum is to increase the sensitivity  
of comparison accuracy. To illustrate, let's consider discretization of a single day of 24 hours with 96 intervals of 15 minutes and a single change point at 8 AM (i.e., 32nd interval). A 30 minute timing discrepancy between predicted and simulated resource level corresponds to an accuracy of 98\% which hinders the comparison of accuracies with high variability. Instead of using all 96 intervals, by using 20 intervals (2.5 hours) centered at 8 AM, the accuracy would be 90\%. Without apriori imperfect knowledge of the change points, the comparison temporal spectrum cannot be determined. The proposed PD-CPD begins with a wide range of the temporal spectrum (centered at initial DD-CPD estimates of change points), i.e. low sensitivity, 
and then gradually constricts the temporal spectrum to increase sensitivity.  
Accuracy of the incumbent resource level solution (change point) is then found by averaging the accuracies across multiple realizations.

\subsubsection{Identifying Optimal Process-Driven Change Point}

In the last step, we perturb and update change points that are determined by PD-CPD method in order to improve prediction results. After training the NARX neural network and testing with the actual system's data, we obtain an average prediction accuracy indicating degree of agreement between the simulated resource levels and system's actual performance data. Next, the incumbent change points are updated by perturbation (i.e., change times), and simulation, NARX training and testing procedure are repeated to evaluate different change point combinations. 

In small instances, one can conduct an extensive neighborhood search, i.e. run simulations and NARX for assessing fitness for a small set of change point combinations to ascertain the optimal change point. However, for medium to large instances, e.g., multiple change points, such an approach may prove impractical and require an optimization approach. The perturbation-based optimization procedure of PD-CPD approach, belongs to the domain of simulation-based optimization, where at each iteration, given a direction of improvement, the incumbent solution is iterated in the improvement direction by a predetermined amount/step size.
Proposed PD-CPD approach starts this perturbation with the initial change points identified through DD-CPD. At the end of first iteration, a perturbation step is executed where change times of resource levels are perturbed to obtain improvement direction.  A common challenge in using solely the direction of improvement for iterations is the nonlinearity of the objective space which poses risk of convergence to a local optima. Hence, most solution approaches also utilize an exploration step where iterates are allowed to be perturbed to an inferior solution (i.e., exploration step) \cite{RN53}. 

While many of the simulation-based optimization approaches are applicable, we use a modified form of the simulated annealing algorithm in our experiments. Simulated annealing is a probabilistic technique that brings together exploitation (choosing most accuracy improving direction) with exploration that chooses suboptimal moves with a gradually decreasing probability to allow for the algorithm to escape local minima. The tradeoff between exploration and exploitation is cast through a temperature parameter, which yields to pure exploitation as the temperature goes to zero, or a random-walk when temperature is infinite. A common approach is to start with a higher temperature to allow exploration in earlier stages, and to lower it after every iteration to ensure stability towards the end of the algorithm. We refer the interested reader to \cite{RN49, RN50} for variants and uses of simulated annealing. Our rational for choosing simulated annealing was to provide a simple search method to showcase the flexibility of our framework in terms of integrating different simulation-based optimization methods. For a general overview of search algorithms within simulation literature that can be integrated into our framework, we refer the interested reader to  \cite{RN51}. 

\subsection{Solution Algorithm}

In this section, we formally introduce the Process-Driven Change Point Detection algorithm that brings together the set of methods outlined in Sections \ref{Sec:DataDriven} \& B. 
Main objective of the algorithm is to leverage detailed DES models to capture process-driven insights, which are then used to augment the initial change point estimations that are acquired through purely data-driven methods.

\begin{algorithm}
\BlankLine
\KwData{Time Series Data Observations}
\KwResult{Process-Driven Change Point Predictions {$\tau^*$}}
\BlankLine
 \textbf{Stage A:} Data-Driven Change Point Estimation: 
 \begin{itemize}
     \item Convert the observations from raw data into time series of the arrivals and features, ${\alpha} :=  \{\alpha_{t}, \; \forall t \in \mathcal{T} \}$, and ${\psi} := \{\psi_{t}^\ell, \; \forall \ell \in \mathcal{F},\forall t \in \mathcal{T} \}$, respectively. $\mathcal{T}$ and $\mathcal{F}$ indicate the set of times and features.
 \end{itemize}
 \begin{itemize}
     \item Define $\tau_0, \cdots, \tau_m \in \mathcal{M}$ as the ordered change points such that $\tau_i < \tau_j$ for any $i<j$. These change points split the sets $\mathbf{\alpha}$ and $\mathbf{\psi}$ to $m$ segments, whereby the $i^{th}$ segment contains $\{ \alpha_t, \;  \forall t \in \mathcal{T} : \tau_{i-1} < t \leq \tau_i \}$ and $\{ \psi_{i,t}, \;   \forall i \in \mathcal{M},\forall t \in \mathcal{T}: \tau_{i-1} < t \leq \tau_i \}$. ${R_i}$ is denoted as the corresponding resource level.
     \item Given $\psi$, identify the optimal data driven change points $\mathbf{\tau}^*_{DD}:= \arg \min_{\tau,m} \sum_{i = 1}^m [\phi({\psi}: \; \tau_{i-1}< t \leq \tau_i)] + \beta m$, where $\phi$ and $\beta$ denote the deviation measurement, and the penalty for additional change points, respectively. 
 \end{itemize}
 
 \textbf{Stage B:} Process-Driven Change Point Prediction:
 
 \begin{itemize}
 \item Let $k \leftarrow 1$, $\tau^0 \leftarrow \mathbf{\tau}^*_{DD}$, $\varepsilon_0 \leftarrow M$,
 where $M$ is a sufficiently large number. Define $\Omega = \emptyset$ as the set of change point scenarios.

 \While{$k \leq k_{max}$}{
 \begin{itemize}
     \item Randomly select $\tau$, an $n$-neighbor of $\tau^{k-1}$, \\
     from the list $\delta(\tau_{k-1})$ that includes all change point combinations $\tau \in \{ \tau_i \; \forall i \in \mathcal{K} : |\tau_i-\tau_i^k| \leq n \}$.
     \item Execute the Procedure $PDA(\mathbf{\alpha},\mathbf{R},\mathbf{\tau}, \mathbf{\psi})$ to obtain \\ the error term associated with $\tau$, namely $\varepsilon$.
     \item If $\varepsilon \geq \varepsilon^{k-1}$, with probability $\exp(\varepsilon - \varepsilon^{k-1}/(k \cdot Temp(k))$, let $\tau^k\leftarrow\tau^{k-1}$ and $\varepsilon^{k} \leftarrow \varepsilon^{{k-1}}$. \\
     Otherwise, $\tau^k \leftarrow \tau$ and $\varepsilon^{{k}} \leftarrow \varepsilon$.
 \item Add the change points $\tau^k$ and the associated error $\epsilon^k$ \\
 to the set $\Omega$. Update $Temp(k)$ and $k \leftarrow k+1$.
 \end{itemize}
 }
 
 \item Optimal change point set is $\tau^* \leftarrow \tau^j$, where $j$ yields minimum average error, i.e. $j = \underset{{1\leq i \leq k}}{\arg \min} \;{\varepsilon^{i}}$.

\end{itemize}
 \caption{Process-Driven Change Point Detection}
\end{algorithm}

\begin{procedure}
\BlankLine
\KwIn{$\mathbf{\alpha},\mathbf{R},\mathbf{\tau}, \mathbf{\psi}$}
\BlankLine
\begin{itemize}
\item
Build simulation model $\xi(\mathbf{\alpha},\mathbf{R},\mathbf{\tau})$ using arrival data $\mathbf{\alpha}$, resource levels $\mathbf{R}$ and change point estimates $\mathbf{\tau}$. \\
Define $\mathcal{S}$ as the set of replications.

\For{$s \in \mathcal{S}$}
{
 \begin{itemize}[leftmargin=*]
 \item Generate a set of time series realizations from \\
 $\xi
 (\mathbf{\alpha},
 \mathbf{R},\mathbf{\tau})$, 
 denote the corresponding feature set as $\psi_s$.
\item Use k-fold cross validation for tuning the parameters \\
of the NARX model. Using the optimized parameters, train NARX model using $\mathbf{\psi}_s$ and $\mathbf{R}$;
 \item Test NARX model using actual feature sets $\mathbf{\psi}$ and $\mathbf{R}$. 
 Denote the corresponding error as $\varepsilon_s$.
 \end{itemize}
}
\item Obtain an average error
 $\varepsilon =
 \sum_{s\in\mathcal{S}}
 \varepsilon_s 
 |\mathcal{S}|$.

 \end{itemize}
 
\KwOut{$\varepsilon$}
\caption{Process-Driven Assessment (PDA)}
\end{procedure}

The proposed algorithm (Algorithm 1) starts with data-driven estimation in Stage A that include acquisition and processing of observation data, and extraction of multi-dimensional time-series features for each observation. These feature sets are used to produce initial, data-driven change point estimates that initialize the search space for Stage B. 
Stage B incorporates process-driven insights to change point prediction. It
starts with selecting change point combinations in the neighborhood of the incumbent solution. For each selected change point combination, multi-replication runs of a DES simulation model is executed as outlined in Section \ref{Sec:SimModel}. The resulting DES simulations are used to train and predict resource levels via NARX models. NARX prediction accuracy measures for observed time-series data streams are evaluated to quantify the success rate of the change point combination. A detailed explanation of this procedure is provided in  \ref{Sec:NARX}, and outlined within the Process-Driven Assessment procedure. Given the accuracy estimates for each candidate change point solution, the algorithm decides which candidate solution replaces the incumbent change point. After sufficient number of iterations, the algorithm terminates, and the change point estimate with minimal error is identified as the optimal prediction. The perturbation based optimization procedure is a modified version of simulated annealing.

\section{Case Study}
To demonstrate the steps of our approach, we conducted a case study using an emergency department (ED) and consider two resource level changes reflecting shift changes within a single day. Resource levels are considered as ${R_i\in }[1,3]$ in our model. This level of resources (along with other parameters and arrival data) corresponds to a realistic ED setting where there are occasional congestion induced waiting. In the pre-processing phase, the input raw data is generated by simulating the ED system for 30 realizations (e.g., 30 day history), which are then converted to time-series data. Since we use snapshot property to generate time series data with $10\:$ minute intervals, each time series input has $144$ data points for $24\:$ hours horizon. In this experiment, we use $6\:$ features (such as the number of entities waiting in the system, waiting in queue, time in idle/busy states) in order to predict resource level change times with multivariate data-driven change point detection. Figure 4 depicts input features and initial change point detection results.

\begin{figure}[h]
    \centering
    \includegraphics[scale=0.28]{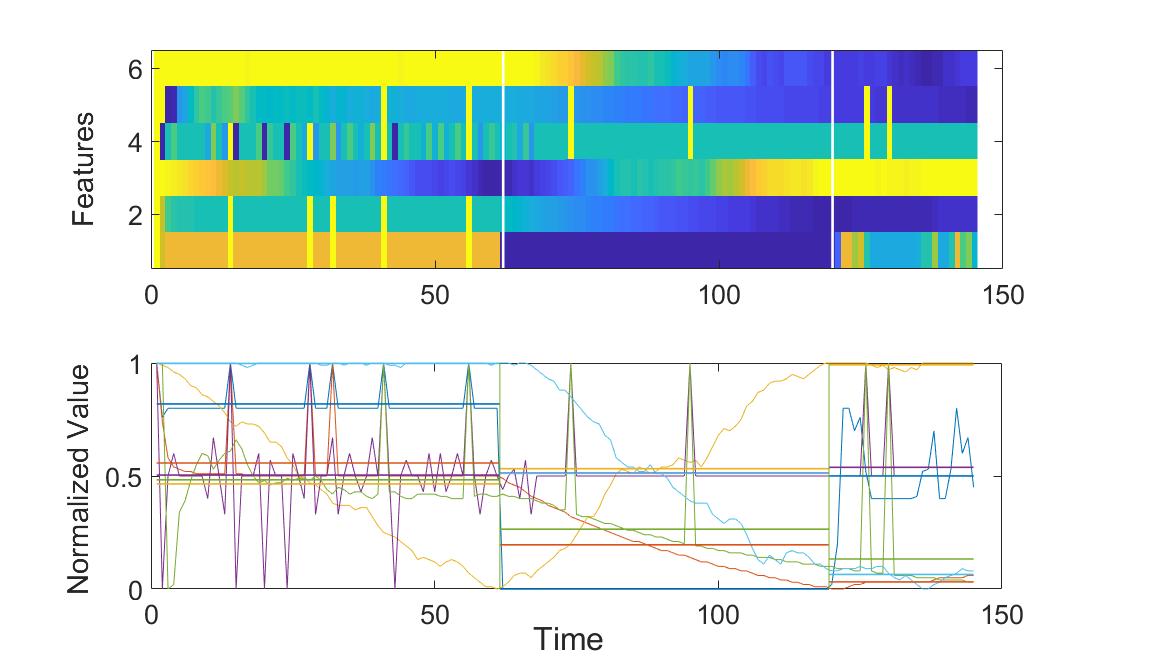}
    \caption{Model Input Features and CPD Results.}
    \label{fig_4}
\end{figure}

In this case study, we use non-parametric offline change point detection method considering "mean" and "standard deviation" changes as the change point detection statistics. For "mean" statistic, change point detection function minimizes the total residual error from the best horizontal level for each section. Given a time series data as $x_1,\;x_2,\;...,\;x_n$, CPD function finds change point $t$ such that:
\begin{align}
    Z(t) =& \sum_{i=1}^{t-1}(x_i-x_1^{t-1})^2 + \sum_{i=t}^{n}(x_i-x_t^{n})^2\\
    = &\sum_{i=1}^{t-1}(x_i-\dfrac{1}{t-1}\sum_{j=1}^{t-1}x_j)^2+\sum_{i=t}^{n}(x_i-\dfrac{1}{n-t+1}\sum_{j=t}^{n}x_j)^2\\
    = &(t-1)\:var([x_1,\:...\:x_{t-1}]) + (n-t+1)\:var([x_t,\:...\:x_n])
\end{align}
attains its minimum value. When we use standard deviation, we fix the mean, and use the following function;
\begin{align}
    &\sum_{i=m}^{n}\varphi(x_i;\:\theta([x_m,\:...\:x_n])) \\
    =&(n-m+1)\log\sum_{i=m}^{n}\sigma^2([x_m,\:...\:x_n]) \\
    =&(n-m+1)\log(\dfrac{1}{n-m+1}\sum_{i=m}^{n}(x_i-\dfrac{1}{n-m+1}\sum_{k=m}^{n}(x_k)^2) \\
    =&(n-m+1)\log var([x_m,\:...\:x_n]).
\end{align}

Data-driven change point detection algorithm identifies two change points of the system as $\tau_1 = 10.5$ and $\tau_2 = 19.33$ during the 24-hr day. Next, we parametrize the discrete event simulation model by setting up the changes to occur at these initial estimates. Next we simulate the ED for each realization with multiple replications and process the outputs of each realization to obtain the multivariate time-series features. 


Next, using simulation outputs, we train and validate a NARX neural network model for each realization. In the training phase, we use multiple inputs (i.e., 6 features representing the performance outputs of the simulated system) and single output (resource level at each time series point) with training samples corresponding to the simulation replications. We use scaled conjugate gradient backpropagation network function for training and validation. We note that the simulation data is only used for the training and validation, not for testing. In the testing phase, actual historical time-series data is used as input features and simulation labels (resource levels) are used for output, thereby we observe simulation model's fitness with respect to the actual data. In this experiment, input delays, feedback delays and hidden layer size are considered as  1:2, 1:2, and 5, respectively. Training and validation sets are divided as $80\%$ and $20\%$, respectively, and testing data is taken from the actual data. 
Figure 5 illustrates the training and validation results for a single NARX neural network. We have two change points and three different resource levels in a day. Since resource shift at $CP_1$ is larger, prediction performance in $\tau_1$ is better than $\tau_2$. We note that accuracy results (both in validation and testing) vary across individual NARX models corresponding to each realization (see Figure 7).

\begin{figure}[h]
    \centering
    \includegraphics[scale=0.28]{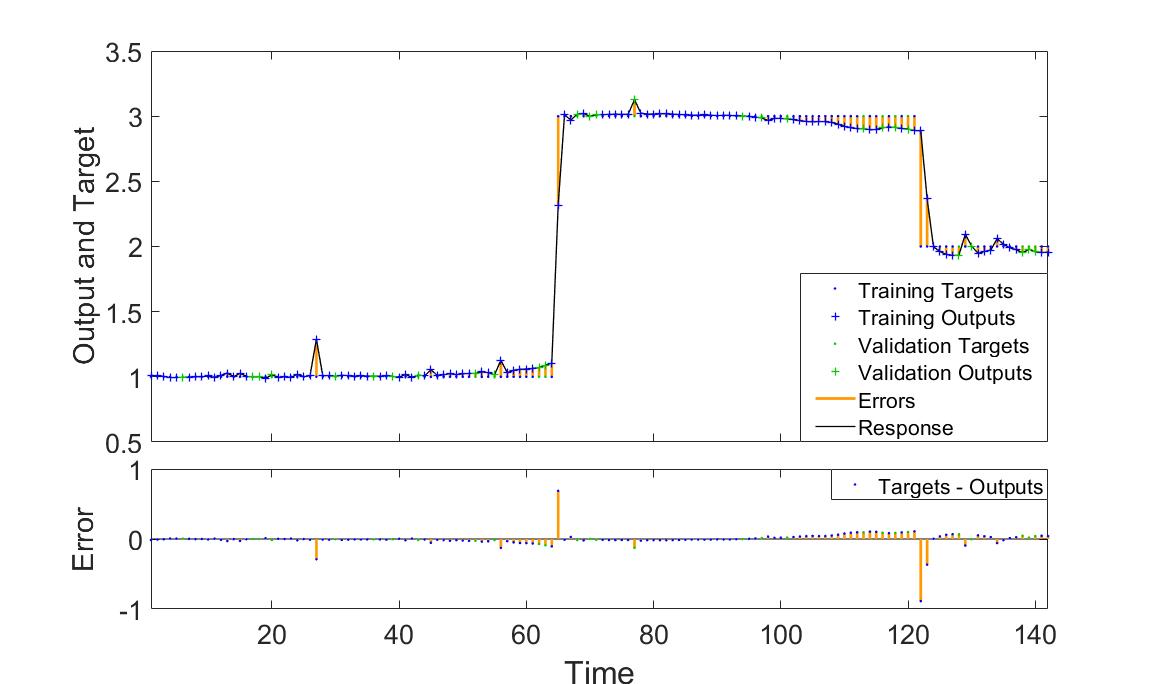}
    \caption{NARX Neural Network Time Series Response for Training and Validation Sets.}
    \label{fig_5}
\end{figure}

For each realization, we test the corresponding NARX neural network model and find the average error across all realizations. Next, we perturb the incumbent change points as described in Algorithm 1. We use a modified version of simulated annealing to perturb the change points and obtain the change point combination that yields the minimum error. The perturbation algorithm converges the CP iterates to the actual CPs of $\tau_1^*=10$ and $\tau_2^*=20$ hours. 

Figure 6 characterizes the absolute time deviation (averaged across realizations) response as a function of different change point combinations used in simulations. We note that our proposed PD-CPD is an unsupervised approach seeking to minimize the deviation between the change points of resource levels that are simulated versus those predicted by the NARX neural network models. Figure 6's depiction thus illustrates how different CPs used in simulations compare with the actual CPs. Absolute time deviation is the absolute distance between the actual CPs ($\tau_1^*=10$ and $\tau_2^*=20$ hours) and those predicted by the NARX models in the testing stage, respectively. Response surface depicted in Figure 6 reveals the non-convexity of the average absolute time deviation in terms of simulated CPs. It further shows that the local minimum is attained at the actual CPs, i.e. $\tau_1^*$ and $\tau_2^*$. The reason for non-zero average time deviation at the minimum is that NARX models for some of the realizations predict change
\begin{figure}[h!]
    \centering
    \includegraphics[scale=0.28]{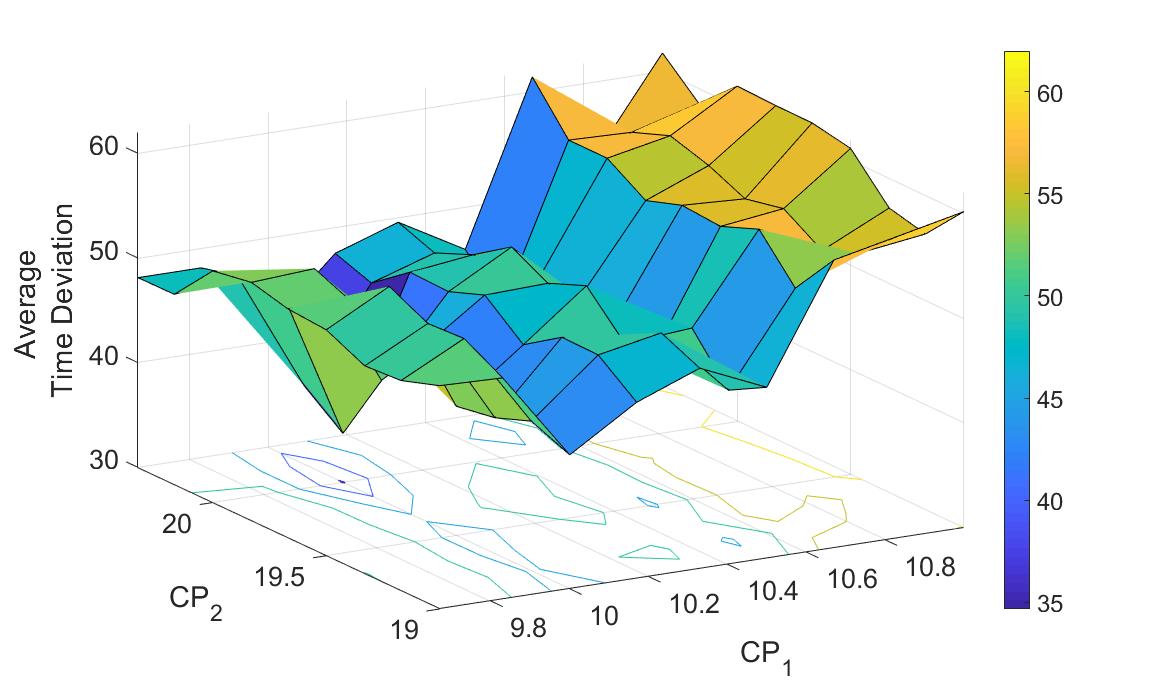}
    \caption{PD-CPD Average Absolute Time Deviation Results.}
    \label{fig_6}
\end{figure}
points different than actual CPs given the time-series features of different realizations. However, the method converges to the  minimizing solution of $\tau_1^*=10$ and $\tau_2^*=20$ hours which has zero deviation from the actual CPs. This is an improvement of the DD-CPD results which has a total absolute deviation of 70 minutes.

Figure 7 compares the results of the DD-CPD and PD-CPD approaches on a daily (realization) basis for 30 days using the optimal solution. Of the 30 days, NARX models' predictions for days 3, 6, 16, and 23 correspond to the actual CPs, i.e. no deviation. The PD-CPD outperforms the DD-CPD in 26 days out of 30 days. Note that while on days 4,26, and 28, PD-CPD's deviations are higher than DD-CPD, these results are for individual days. Indeed, the PD-CPD's optimizing solution corresponds to actual CPs which are better than the DD-CPD results of $\tau_1 = 10.5$ and $\tau_2 = 19.33$ which has a total absolute deviation of 70 minutes. 

\begin{figure}[h]
    \centering
    \includegraphics[scale=0.20]{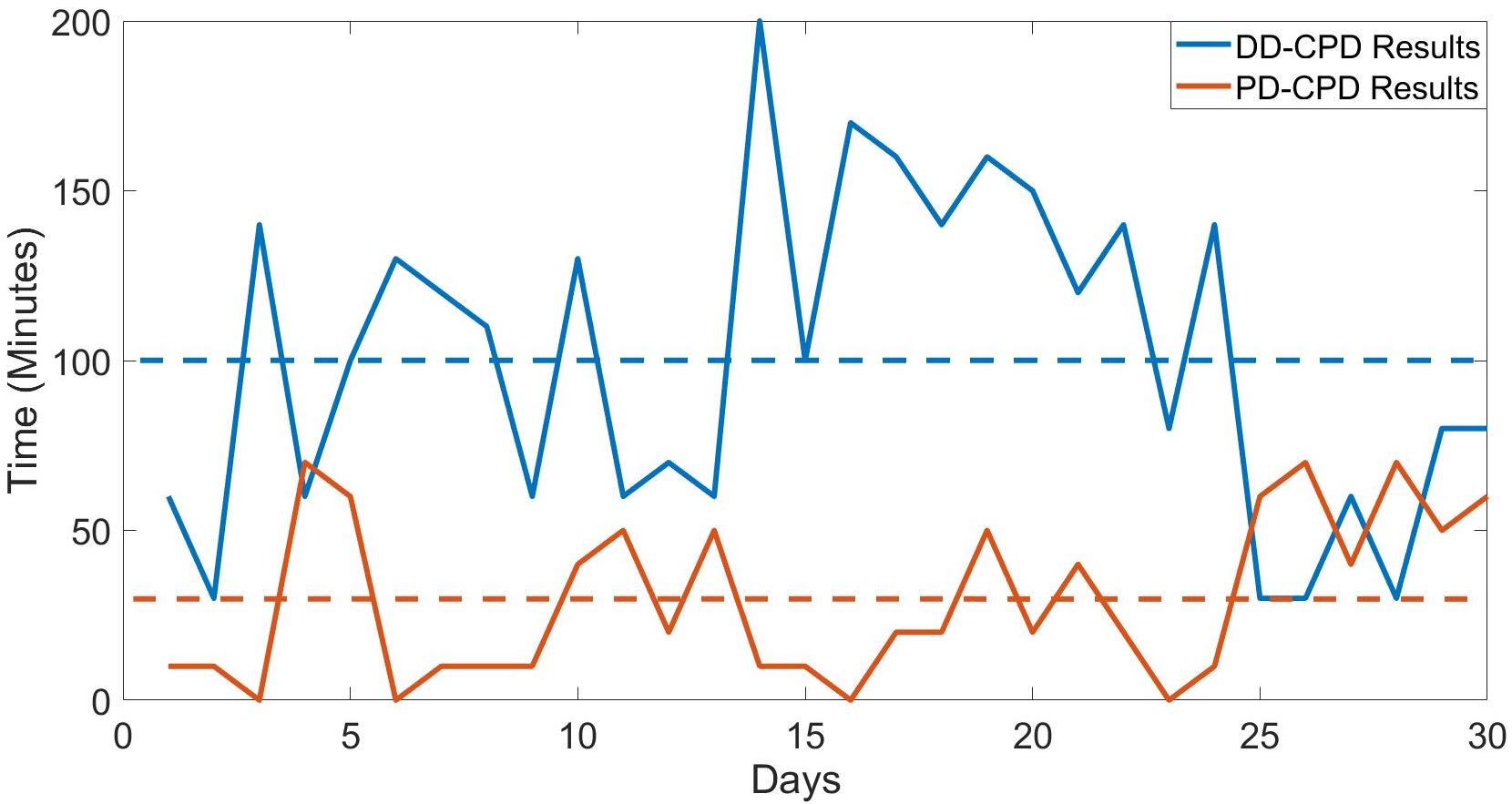}
    \caption{DD-CPD vs PD-CPD - Daily Based Absolute Time Deviation Results.}
    \label{fig_7}
\end{figure}

\section{Limitation and Future Directions}
In the current study, each day is simulated with multiple replications and a NARX model is trained based on the average features across replications. Instead of averaging, one approach could be to train and validate NARX model with the non-averaged replication results. However, NARX model is not able to process more than one time series sequence as it is not able to handle the time delay between observations across a day’s replications. An alternative way (to averaging across replications) is to replicate each day’s simulation and build a NARX model for each of these simulation replications. These NARX models would then form an ensemble model for each day. We can then make aggregate predictions using these ensemble models, i.e. by choosing the majority resource level  across simulation replications.

One future research opportunity is to extend the present approach to change point detection at multiple timescales. Most discrete event systems have periodic change points that are short term(i.e. daily, weekly) as well as long term (i.e. monthly, quarterly, or yearly). This study focused on detecting change points that repeat on a single time scale where we considered daily shift changes. Future extension can investigate the presence of short-term and long-term changes (i.e., quarterly surgery block time allocation and daily staff shift changes) and their joint detection with process knowledge driven CPD method.


\section{Conclusion}
In this paper, we propose a novel change point detection algorithm that combines data driven methods with process knowledge, to identify when resource levels change in discrete event systems. Unique to our approach is the use of discrete event simulation models to capture complex process dynamics that typically occur at times of resource level transition. Our experimental results indicate that the value of process knowledge is significant in improving change point detection accuracy. The proposed model can complement a large variety of data driven change point detection models, 
and provides an extensive basis for automated discovery of process knowledge in discrete event systems.




\bibliographystyle{IEEEtran}
\bibliography{references.bib}



%






\end{document}